\documentclass[conference]{IEEEtran}
\usepackage{algorithm2e}
\usepackage{algorithmic}
\usepackage{url}
\usepackage{graphicx}

\begin{document}

\title{Temporally Coherent Bayesian Models for Entity Discovery in Videos by Tracklet Clustering}

\maketitle

\maketitle
\begin{abstract}
A video can be represented as a sequence of tracklets, each spanning 10-20 frames, and associated with one entity (eg. a person).  The task of \emph{Entity Discovery} in 
videos  can be naturally posed as tracklet clustering. We approach this task by leveraging \emph{Temporal Coherence}(TC): the fundamental property of videos that each tracklet is 
likely to be associated with the same entity as its temporal neighbors. Our major contributions are the first Bayesian nonparametric models for TC at tracklet-level. We extend 
Chinese Restaurant Process (CRP) to propose TC-CRP, and further to Temporally Coherent Chinese Restaurant Franchise (TC-CRF) to jointly model short temporal segments. On the task of 
discovering persons in TV serial videos without meta-data like scripts, these methods show considerable improvement in cluster purity and person coverage compared to state-of-the-art 
approaches to tracklet clustering. We represent entities with mixture components, and tracklets with vectors of very generic features, which can work for any type of entity (not 
necessarily person). The proposed methods can perform online tracklet clustering on streaming videos with little performance deterioration unlike existing approaches, and can 
automatically reject tracklets resulting from false detections. Finally we discuss entity-driven video summarization- where some temporal segments of the video are selected 
automatically based on the discovered entities.
\end{abstract}

\IEEEpeerreviewmaketitle

\section{Introduction}
Online video repositories like Youtube, Dailymotion etc have been experiencing an explosion of user-generated videos. Such videos are often shot/recorded from the television by 
users, and uploaded onto these sites. They have very little metadata like dialogue scripts, or a textual summary/representation of the content. When an user searches these 
repositories by keywords, (s)he is suggested hundreds of videos, out of which (s)he may choose a small number for viewing. This has given rise to the topic of \emph{Video 
Summarization}~\cite{vsumm}, which aims to provide the user a short but comprehensive \emph{summary} of the video. However, the current state-of-the-art mostly provides a few 
keyframes as summary, which may not have much semantic significance. The high-level semantic information of videos that is most important to users is carried by \emph{entities}- 
such as persons or other objects. With the recent progress in object detection~\cite{felzen}~\cite{facedet} in single images and videos, it is now possible to have a high-level 
representation of videos in terms of such entities. One effective way of summarization is to have a list of entities that appear frequently in a video. Further, an user may want 
to watch only a part of a video, for example wherever a particular person (or set of persons) appears, which motivate the tasks of Entity Discovery and Entity-driven Summarization
~\cite{csum} of videos.

The problem of \emph{automated discovery of persons from videos along with all their occurrences} has attracted a lot of interest~\cite{castlist}\cite{scr1}\cite{scr2} in video 
analytics. 
Existing attempts try to leverage meta-data such as scripts~\cite{scr1}\cite{scr2} and hence do not apply to videos available on the wild, such as TV-Series episodes uploaded by 
viewers on Youtube (which have no such meta-data). In this paper, we pose this problem as \emph{tracklet clustering}, as done in~\cite{trklnk}. Our goal is to design algorithms 
for tracklet clustering which can work on long videos. Tracklets\cite{tracklet} are formed by detections of an entity (say a person) from a short contiguous sequence of 10-20 
video frames. They have complex spatio-temporal properties. We should be able to handle any type of entity, not just person.
Given a video in the wild it is unlikely that the number of entities will be known, so the method should automatically adapt to unknown number of 
entities. To this end we advocate a \emph{Bayesian non-parametric} clustering approach to Tracklet clustering and study its effectiveness in automated discovery of entities 
with all their occurrences in long videos. The main challenges are in modeling the spatio-temporal properties. To the best of our knowledge this problem has not been studied 
either in Machine Learning or in Computer Vision community.

\begin{figure}
 \centering
\includegraphics[width=3.3in,height=1.2in]{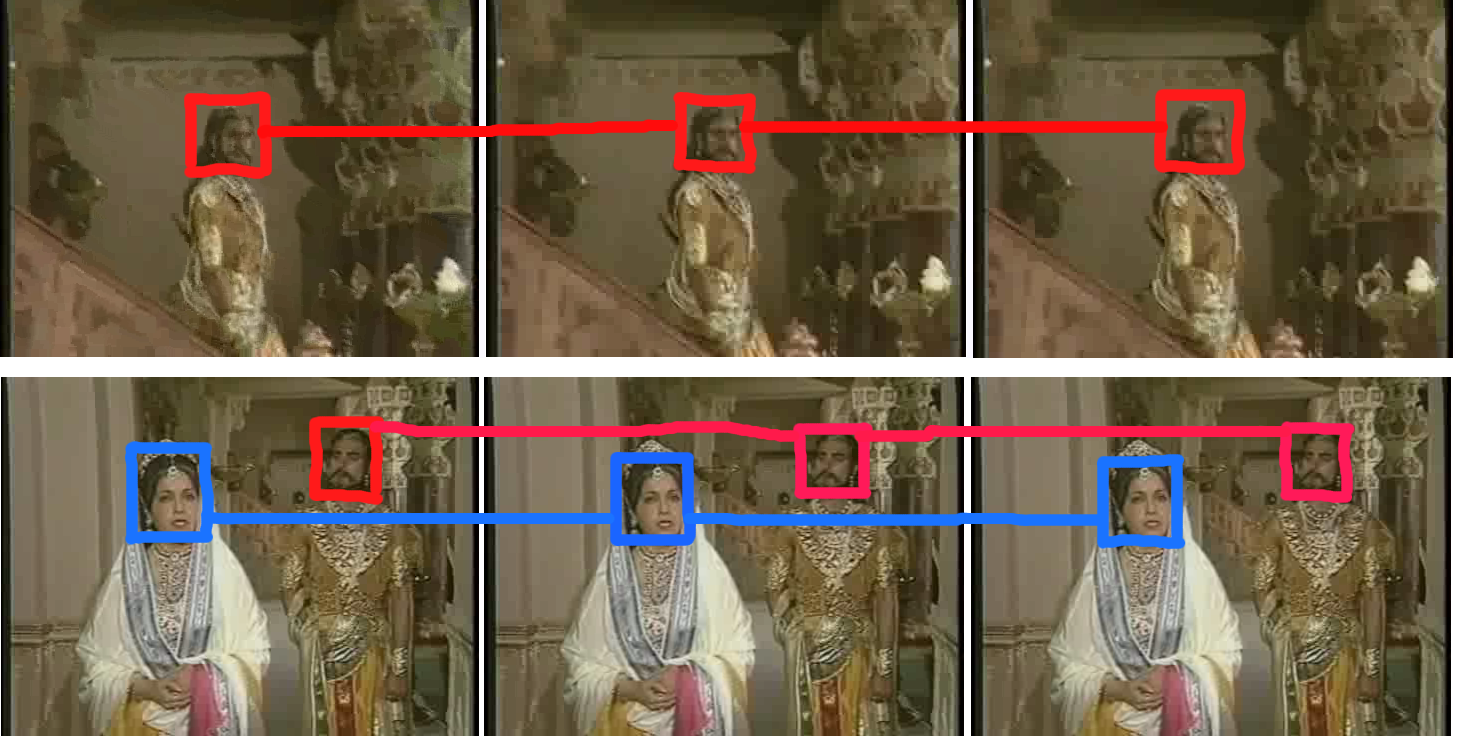} 
\caption{\small{Top: a window consisting of frames 20000,20001,20002, Bottom: another window- with frames 21000,21001,21002. The detections 
are linked on spatio-temporal basis to form tracklets. One person (marked with red) occurs in both windows, the other 
character (marked with blue) occurs only in the second. The two red tracklets should be associated though they are from non-contiguous windows}}
\label{fig:asso}
\end{figure}

To explain the spatio-temporal properties we introduce some definitions.
A \emph{track} is formed by detecting entities (like people's faces) in each video frame, and associating detections across a contiguous sequence of frames (typically a few 
hundreds in a TV series) based on \emph{appearance} and \emph{spatio-temporal} locality. Each track corresponds to a particular entity, like a person in a TV series. 
Forming long tracks is often difficult, especially if there are multiple detections per frame. This can be solved hierarchically, by associating the detections in a short 
window of frames (typically 10-20) to form \emph{tracklets}~\cite{tracklet} and then linking the tracklets from successive windows to form tracks.
The \emph{short-range association of tracklets} to form tracks is known as \emph{tracking}. But in a TV series video, the same person may appear in different (non-contiguous) 
parts of the video, and so we need to associate tracklets on a \emph{long-range} basis also (see Figure~\ref{fig:asso}). Moreover the task is complicated by lots of 
\emph{false detections} which act as spoilers. Finally, the task becomes more difficult on streaming videos, where only one pass is possible over the sequence.

\begin{figure}
 \centering
\includegraphics[width=3.3in,height=0.35in]{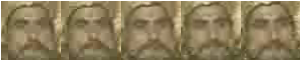} 
\caption{\small{TC at Detection level: Detections in successive frames (linked to form a tracklet) are almost identical in appearance, i.e. have nearly identical
 visual features}}
\includegraphics[width=3.3in,height=0.4in]{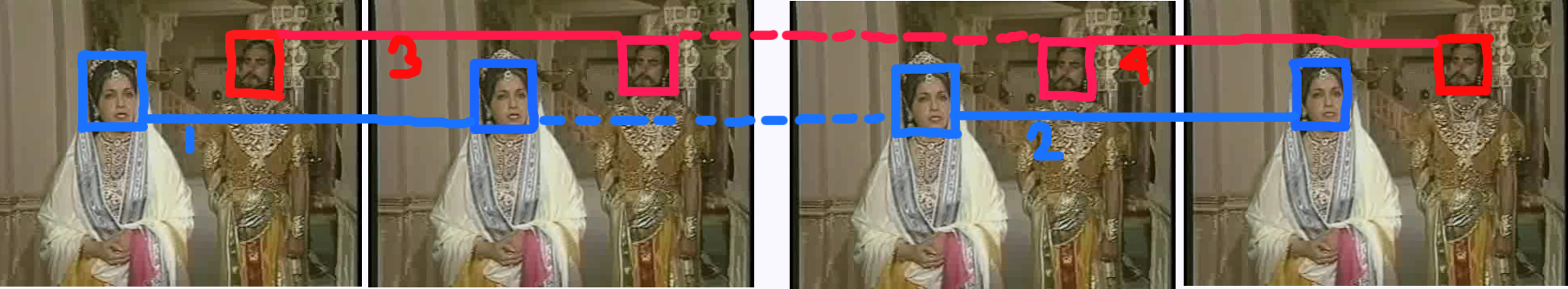} 
\caption{\small{TC at Tracklet level: Blue tracklets 1,2 are spatio-temporally close (connected by broken lines), and belong to the same person. Similarly red 
tracklets 3 and 4.}}
\label{fig:tc}
\end{figure}

A major cue for this task comes from a very fundamental property of videos: \emph{Temporal Coherence}(TC). This property manifests itself at detection-level as well as 
tracklet-level; at feature-level as well as at semantic-level. At detection-level this property implies that the visual features of the detections (eg. appearance of an entity) 
are almost unchanged across a tracklet (See Fig. 2). At tracklet-level it implies that \emph{spatio-temporally close (but non-overlapping) tracklets are likely to 
belong to the same entity} (Fig.~\ref{fig:tc}). Additionally, \emph{overlapping tracklets (that span the same frames), cannot belong to the same entity}.
A tracklet can be easily represented as all the associated detections are very similar (due to detection-level TC). Such representation is not easy for a long track where the 
appearances of the detections may gradually change.

\textbf{Contribution}
Broadly, this paper has two major contributions: it presents the first Bayesian nonparametric models for TC in videos, and also the first entity-driven approach to video 
modelling. To these ends, we explore tracklet clustering, an active area of research in Computer Vision, and advocate a Bayesian non-parametric(BNP) approach for it. 
We apply it to an important open problem: discovering entities (like persons) and all their occurrences from long videos, in absence of any meta-data, e.g. scripts.
We use a simple and generic representation leading to representing a video by a matrix, whose columns represent individual tracklets (unlike other works which represent an 
individual detection by a matrix column, and then try to encode the tracklet membership information). We propose Temporally Coherent-Chinese Restaurant process(TC-CRP), a BNP 
prior for enforcing coherence on the tracklets. Our method yields a superior clustering of tracklets over several baselines especially on long videos.
As an advantage it does not need the number of clusters in advance. It is also able to automatically filter out false detections, and perform the same task on \emph{streaming 
videos}, which are impossible for existing methods of tracklet clustering. We extend TC-CRP to the Temporally Coherent Chinese Restaurant Franchise (TC-CRF), that jointly models 
short video segments and further improves the results. We show that the proposed methods can be applied to entity-driven video summarization, by selecting a few 
representative segments of the video in terms of the discovered entities.

\section{Problem Definition}\label{sec:def}
In this section, we elaborate on our task of tracklet clustering for entity discovery in videos. 

\subsection{Notation}
In this work, given a video, we fix beforehand the \emph{type of entity} (eg. person/face, cars, planes, trees) we are interested in, and choose an appropriate detector
 like~\cite{facedet}~\cite{felzen}, which is run on every frame of the input video.
The detections in successive frames are then linked based on spatial locality, to obtain 
tracklets. At most $R$ detections from $R$ contiguous frames are linked like this. The tracklets of length less than $R$ are discarded, hence all tracklets consist of $R$ 
detections. 
We restrict the length of tracklets so that the appearance of the detections remain almost unchanged (due to detection-level TC), which facilitates tracklet representation. 
At $R=1$ we work with the individual detections.

We represent a detection by a vector of dimension $d$. This can be done by downscaling a rectangular detection to $d\times d$ square and then reshaping it to 
a $d^2$-dimensional vector of pixel intensity values (or some other features if deemed appropriate). Each tracklet $i$ is a collection of $R$ detections $\{I^i_1,\dots,I^i_R\}$.
 Let the tracklet $i$ be represented by $Y_i=\frac{\sum_{j=1}^R{I^i_j}}{R}$. So finally we have $N$ vectors ($N$: number of tracklets).

The tracklets can be sorted topologically based on their starting and ending frame indices. Each tracklet $i$ has a \emph{predecessor tracklet} $prev(i)$ and a 
\emph{successor tracklet} $next(i)$.
Also each tracklet $i$ has a conflicting set of tracklets $CF(i)$ which span frame(s) that overlap with the frames spanned by $i$. 
Each detection (and tracklet) is associated with an entity, which are unknown in number, but presumably much less than the number of detections (and tracklets). 
These entities also are represented by vectors, say $\phi_1,\phi_2,\dots,\phi_K$. Each tracklet $i$ is associated with an entity indexed by $Z_i$, i.e. $Z_i \in \{1,2,\dots,K\}$.

\subsection{Entity Discovery}
Let each video be represented as a sequence of $d$-dimensional vectors $\{Y_1,\dots,Y_N\}$ along with the set $\{prev(i),next(i),CF(i)\}_{i=1}^N$.
We aim to learn the vectors $\{\phi_1,\phi_2,\dots,\}$ and the assignment variables $\{Z_i\}_{i=1}^N$. In addition, we have \emph{constraints}
 arising out of \emph{temporal coherence} and other properties of videos. Each tracklet $i$ is likely to be associated with the entity that its predecessor or successor is
associated with, except at shot/scene changepoints. Moreover, a tracklet $i$ cannot share an entity with its conflicting tracklets $CF(i)$, as the 
same entity cannot occur twice in the same frame. This notion is considered in relevant literature~\cite{wbslrr}~\cite{fcclus}. Mathematically, the constraints are:
\begin{eqnarray}\label{eq:cons}
 Z_{prev(i)}=Z_i=Z_{next(i)} \forall i \in \{1,\dots,N\} \textit{  w.h.p.} \nonumber \\
 Z_i \notin \{Z_j : j \in CF(i)\} \forall i \in \{1,\dots,N\}
\end{eqnarray}
\emph{Learning a $\phi_k$-vector is equivalent to discovering an entity, and its associated tracklets are discovered by learning the set $\{i: Z(i)=k\}$.} 
These constraints give the task a flavour of non-parametric constrained clustering with must-link and don't-link constraints.

Finally, the video frames can be grouped into short segments, based on the starting frame numbers $F(1),F(2),\dots,F(N)$ of the $N$ tracklets. Consider two successive tracklets 
$i$ and $(i+1)$, with starting 
frames $F(i)$ and $F(i+1)$. If the gap between frames $F(i)$ and $F(i+1)$ is larger than some threshold, then we consider a new temporal segment 
of the video starting from $F(i+1)$, and add $i+1$ to a list of \emph{changepoints} (CP). The beginning of a new temporal segment does not necessarily mean a scene change, the 
large gap between frames $F(i)$ and $F(i+1)$ may be caused by failure of detection or tracklet creation. The segment index of each tracklet $i$ is denoted by $S(i)$.

\section{Related Work}
\textbf{Person Discovery in Videos} is a task which has recently received attention in Computer Vision. Cast Listing~\cite{castlist} is aimed to choose a representative 
subset of the face detections or face tracks in a movie/TV series episode. Another task is to label \emph{all the detections} in a video, but this requires movie scripts
~\cite{scr1} or labelled training videos having the same characters~\cite{scr2}. Scene segmentation and person discovery are done simultaneously using a generative model in 
~\cite{vidpars}, but once again with the help of scripts. An unsupervised version of this task is considered in~\cite{fcclus}, which performs 
\textbf{face clustering} in presence of spatio-temporal constraints as already discussed. For this purpose they use a Markov Random Field, and encode the constraints as clique 
potentials. Another recent approach to face clustering is~\cite{wbslrr} which incorporates some spatio-temporal constraints into subspace clustering.

\textbf{Tracklet Association} \emph{Tracking} is a core topic in computer vision, in which a target object is located in each frame based on appearance similarity and spatio-
temporal locality. A more advanced task is \emph{multi-target tracking}~\cite{track}, in which several targets are present per frame. A tracking 
paradigm that is particularly helpful in multi-target tracking is \emph{tracking by detection}~\cite{dettrack}, where object-specific detectors like~\cite{felzen} are run per 
frame (or on a subset of frames), and the detection responses are linked to form tracks. From this came the concept of \emph{tracklet}~\cite{tracklet} which attempts to do the 
linking hierarchically. This requires pairwise similarity measures between tracklets. Multi-target tracking via tracklets is usually cast as Bipartite Matching, which is 
solved using Hungarian Algorithm.  Tracklet association and face clustering are done simultaneously in~\cite{trklnk} using HMRF. 
The main difference of face/tracklet clustering and person discovery is that, the number of clusters to be formed is not known in the latter.

Independent of videos, \textbf{Constrained Clustering} is itself a field of research. Constraints are usually \emph{must-link and don't-link}, which specify pairs which should 
be assigned the same cluster, or must not be assigned the same cluster. A detailed survey is found in~\cite{ccsurvey}. The constraints can be hard~\cite{ckmeans} or 
soft/probabilistic~\cite{ppc}. Constrained Spectral Clustering has also been studied recently~\cite{conspec1}~\cite{conspec2}, which allow constrained clustering of datapoints 
based on arbitrary similarity measures.

All the above methods suffer from a major defect- the number of clusters needs to be known beforehand. A way to avoid this is provided by \textbf{Dirichlet Process}, which is 
able to identify the number of clusters from the data. It is a mixture model with infinite number of mixture components, and each datapoint is assigned to one component.
A limitation of DP is that it is exchangeable, and cannot capture sequential structure in the data. For this 
purpose, a Markovian variation was proposed: Hierarchical Dirichlet Process- Hidden Markov Model (HDP-HMM). A variant of this is the \emph{sticky} HDP-HMM 
(sHDP-HMM)~\cite{hdphmm}, which was proposed for temporal coherence in speech data for the task of speaker diarization, based on the observation that successive datapoints 
are likely to be from the same speaker and so should be assigned to the same component. Another Bayesian nonparametric approach for sequential data is the Distance-Dependent 
Chinese Restaurant Process (DDCRP)~\cite{ddcrp}, which defines distances between every pair of datapoints, and each point is linked to another with probability proportional to 
such distances. A BNP model for subset selection is Indian Buffet Process (IBP)~\cite{IBP}, a generative process for a sequence of binary vectors. This has been used for selecting
 a sparse subset of mixture components (topics) in Focussed Topic Modelling~\cite{FTM} as the Compound Dirichlet Mixture Model.

Finally, \textbf{Video Summarization} has been studied for a few years in the Computer Vision community. The aim is to provide a short but comprehensive summary of videos. This 
summary is usually in the form of a few \emph{keyframes}, and sometimes as a short segment of the video around these keyframes. A recent example is ~\cite{vsumm} which 
models a video as a matrix, each frame as a column, and each keyframe as a \emph{basis vector}, in terms of which the other columns are expressed. A more recent work~\cite{KTS} 
considers a kernel matrix to encode similarities between pairs of frames, uses it for 
\emph{Temporal Segmentation} of the video, assigns an importance label to each of these segments using an SVM (trained from segmented and labelled videos), and creates 
the summary with the important segments. However, such summaries are in terms of low-level visual features, rather than high-level semantic features which humans use. An 
attempt to bridge this gap was made in~\cite{csum}, which defined movie scenes and summaries in terms of characters. This work used face detections along with 
\emph{movie scripts} for semantic segmentation into shots and scenes, which were used for summarization.

\section{Generative Process for Tracklets}\label{sec:tccrp}
We now explain our Bayesian Nonparametric model TC-CRP to handle the spatio-temporal constraints (Eq~\ref{eq:cons}) for tracklet clustering, and describe a generative process for
 videos based on tracklets.
\subsection{Bayesian Nonparametric Modelling}
In Section~\ref{sec:def}, we discussed the vectors $\phi_1,\phi_2,\dots$ each of which represent an entity. In this paper we consider a Bayesian approach with Gaussian 
Mixture components $\mathcal{N}(\phi_k,\Sigma_1)$ to account for the variations in visual features of the detections, say face detections of a person. As already mentioned, number of
 components $K$ is not known beforehand, and must be discovered from the data. That is why we consider nonparametric Bayesian modelling. Also, as we shall see, this route allows 
us to elegantly model the temporal coherence constraints. In this approach, we shall represent entities as mixture components and tracklets as draws from such mixture components.

Dirichlet Process~\cite{DP} has become an important clustering tool in recent years. Its greatest strength is that unlike K-means, it is able to discover the correct number of 
clusters. Dirichlet Process is a distribution over distributions over a measurable space. A discrete distribution $P$ is said to be distributed as $DP(\alpha,H)$ over space $A$ 
if for every finite partition of $A$ as $\{A_1,A_2,\dots,A_K\}$, the quantity $\{P(A_1),\dots,P(A_K)\}$ is distributed as $Dirichlet(\alpha H(A_1),\dots,\alpha H(A_K))$, where 
$\alpha$ is a scalar called \emph{concentration parameter}, and $H$ is a distribution over $A$ called Base Distribution. A distribution $P \sim DP(\alpha,H)$ is a discrete 
distribution, with infinite support set $\{\phi_k\}$, which are draws from $H$, called \emph{atoms}.

\subsection{Modeling Tracklets by Dirichlet Process}
We consider $H$ to be a $d-dimensional$ multivariate Gaussian with parameters $\mu$ and $\Sigma_0$. Each atom corresponds to an entity (eg. a person). 
The generative process for the set $\{Y_i\}_{i=1}^N$ is then as follows:
\begin{small}
\begin{equation}\label{eq:dp1}
 P \sim DP(\alpha,H);  X_i \sim P, Y_i \sim \mathcal{N}(X_i,\Sigma_1) \forall i \in [1,N]    
\end{equation}
\end{small}
Here $X_i$ is an atom. $Y_i$ is a tracklet representation corresponding to the entity, and its slight variation from $X_i$ 
(due to effects like lighting and pose variation) is modelled using $\mathcal{N}(X_i,\Sigma_1)$.

Using the constructive definition of Dirichlet Process, called the Stick-Breaking Process~\cite{sethu}, the above process can also be written equivalently as
\begin{small}
\begin{eqnarray}\label{eq:dp2}
 \hat{\pi}_k \sim Beta(1,\alpha), \pi_k= \hat{\pi}_k\prod_{i=1}^{k-1}{(1-\hat{\pi}_{i-1})}, \phi_k \sim H \textbf{     }\forall k \in [1,\infty) \nonumber \\
 Z_i \sim \pi, Y_i \sim \mathcal{N}(\phi_{Z_i},\Sigma_1)  \forall i \in [1,N]
\end{eqnarray}
\end{small}
Here $\pi$ is a distribution over integers, and $Z_i$ is an integer that indexes the component corresponding to the tracklet $i$.
Our aim is to discover the values $\phi_k$, which will give us the entities, and also to find the values $\{Z_i\}$, which define a clustering of the tracklets. 
For this purpose we use collapsed Gibbs Sampling, where we integrate out the $P$ in Equation~\ref{eq:dp1} or $G$ in Equation~\ref{eq:dp2}. The Gibbs Sampling 
Equations $p(Z_i|Z_{-i},\{\phi_k\},Y)$ and $p(\phi_k|\phi_{-k},Z,Y)$ are given in~\cite{gdp}. For $Z_i$,
\begin{small}
\begin{equation}
  p(Z_i=k|Z_{-i},\phi_k, Y_i) \propto p(Z_i=k|Z_{-i})p(Y_i|Z_i=k,\phi)
\end{equation}
\end{small}
Here, $p(Y_i|Z_i=k,\phi)=\mathcal{N}(Y_i|\phi_{k},\Sigma_1)$ is the data likelihood term. We focus on the part $p(Z_i=k|Z_{-i})$ to model TC.

\subsection{Temporally Coherent Chinese Restaurant Process}
In the generative process (Equation~\ref{eq:dp2}) all the $Z_i$ are drawn IID conditioned on $\pi$. Such models are called \emph{Completely Exchangeable}. 
This is, however, often not a good idea for sequential data such as videos. In Markovian Models like sticky HDP-HMM, $Z_i$ is drawn conditioned on $\pi$ and $Z_{i-1}$. 
In case of DP, the independence among $Z_i$-s is lost on integrating out $\pi$. After integration the generative process of 
Eq~\ref{eq:dp2} can be redefined as 
\begin{small}
\begin{eqnarray}
 \phi_{k} \sim H \forall k ; Z_i|Z_1,\dots,Z_{i-1} \sim CRP(\alpha); Y_i \sim \mathcal{N}(\phi_{Z_i},\Sigma_1)
\end{eqnarray}
\end{small}
The predictive distribution for $Z_i|Z_1,\dots,Z_{i-1}$ for Dirichlet Process is known as Chinese Restaurant Process (CRP). It is defined as
\begin{small}
$ p(Z_i=k|Z_{1:i-1}) = \frac{N_k^i}{N-1+\alpha}  \textbf{ if }  k \in \{Z_1,\dots,Z_{i-1}\}; 
= \frac{\alpha}{N-1+\alpha} \textbf{ otherwise}$
\end{small}
where $N_k^i$ is the number of times the value $k$ is taken in the set $\{Z_1,\dots,Z_{i-1}\}$.

We now modify CRP to handle the Spatio-temporal cues (Eq~\ref{eq:cons}) mentioned in the previous section. In the generative process, we
define $p(Z_i|Z_1,\dots,Z_{i-1})$ with respect to $prev(i)$, similar to the Block Exchangeable Mixture Model as defined in~\cite{BE}. Here, with each $Z_i$ we associate a 
\emph{binary change variable} $C_i$. If $C_i=0$ then $Z_i=Z_{prev(i)}$, i.e the tracklet identity is maintained. But if $C_i=1$, a 
new value of $Z_i$ is sampled. Note that every tracklet $i$ has a temporal predecessor $prev(i)$. However, if this predecessor is spatio-temporally close, then it is more 
likely to have the same label. So, the probability distribution of change variable $C_i$ should depend on this closeness. In TC-CRP, we use two values 
($\kappa_1$ and $\kappa_2$) for the Bernoulli parameter for the change variables. We put a threshold on the spatio-temporal distance between $i$ and $prev(i)$, and choose a 
Bernoulli parameter for $C_i$ based on whether this threshold is exceeded or not. Note that maintaining tracklet identity by setting $C_i=0$ is equivalent to \emph{tracking}.

Several datapoints (tracklets) arise due to false detections. We need a way to model these. Since these are very different from the Base mean $\mu$, we consider a 
separate component $Z=0$ with mean $\mu$ and a very large covariance $\Sigma_2$, which can account for such variations. The Predictive Probability function(PPF) for TC-CRP is 
defined as follows:

\begin{small}
\begin{eqnarray}\label{eq:PPF1}
 T(Z_i=k|Z_{1:i-1},C_{1:i-1},C_i=1) = 0 \textbf{ if } k \in \{Z_{CF(i)}\}-\{0\}& \nonumber \\
                                    \propto \beta  \textbf{ if } k=0 &  \nonumber \\
				    \propto n^{ZC}_{k1}\textbf{ if } k \in \{Z_1,\dots,Z_{i-1}\}, k \notin \{Z_{CF(i)}\}& \nonumber \\
                                    \propto \alpha \textbf{ otherwise}& 
\end{eqnarray}
\end{small}
where $Z_{CF(i)}$ is the set of values of $Z$ for the set of tracklets $CF(i)$ that overlap with $i$, and $n^{ZC}_{k1}$ is the number of points $j$ ($j < i$) where $Z_j=k$ and 
$C_j=1$. The first rule ensures that two overlapping tracklets cannot have same value of $Z$. The second rule accounts for false tracklets. The third and fourth rules define
 a CRP restricted to the changepoints where $C_j=1$. The final tracklet generative process is as follows:
\begin{algorithm}[ht!]
\caption{TC-CRP Tracklet Generative Process}\label{hdp}
\begin{algorithmic}[1]
\footnotesize 
 \STATE $\phi_{k} \sim \mathcal{N}(\mu,\Sigma_0)$ $\forall k \in [1,\infty)$
 \FOR {$i=1:N$}
 \IF {$dist(i,prev(i)) \leq thres$} \STATE $C_i \sim Ber(\kappa1)$
 \ELSE \STATE $C_i \sim Ber(\kappa2)$
 \ENDIF
 \IF {$C_i=1$} \STATE draw $Z_i \sim T(Z_i|Z_1,\dots,Z_{i-1},C_1,\dots,C_{i-1},\alpha)$
 \ELSE \STATE $Z_i=Z_{prev(i)}$
 \ENDIF
 \IF {$Z_i=0$} \STATE $Y_i \sim \mathcal{N}(\mu,\Sigma_2)$
 \ELSE \STATE $Y_i \sim \mathcal{N}(\phi_{Z_i},\Sigma_1)$
 \ENDIF
 \ENDFOR
\end{algorithmic}
\end{algorithm}
where $T$ is the PPF for TC-CRP, defined in Eq~\ref{eq:PPF1}.

\subsection{Inference}
Inference in TC-CRP can be performed easily through Gibbs Sampling. We need to infer $C_i$, $Z_i$ and $\phi_k$. As $C_i$ and $Z_i$ are coupled, we sample them in a block 
for each $i \in [1,N]$ as done in~\cite{BE}. If $C_{i+1}=0$ and $Z_{i+1}\neq Z_{i-1}$, then we must have $C_{i}=1$ and $Z_{i}=Z_{i+1}$. If $C_{i+1}=0$ and $Z_{i+1}=Z_{i}$, 
then $Z_{i}=Z_{i+1}$, and $C_{i}$ is sampled from $Bernoulli(\kappa)$. In case $C_{i+1}=1$ and $Z_{i+1}\neq Z_{i-1}$, then $(C_i=a,Z_{i}=k)$ with probability proportional to 
$p(C_i=a)p(Z_i|Z_{-i},C_{i}=a))p(Y_i|Z_i=k,\phi_k)$. If $a=0$ then $p(Z_i=k|Z_{-i},C_i=1)=1$ if $Z_{i-1}=k$, and $0$ otherwise. If $a=1$ then $p(Z_i|Z_{-i},C_{i}=a))$ is 
governed by TC-CRP. For sampling $\phi_k$, we make use of the Conjugate Prior formula of Gaussians, to obtain the Gaussian posterior with mean 
${(n_k\Sigma_1^{-1}+\Sigma^{-1})}^{-1}(\Sigma_1^{-1}Y_k+\Sigma^{-1}\mu)$ where $n_k=|\{i: Z_i=k\}|$, and $Y_k=\sum_{i:Z_i=k}Y_i$.
Finally, we update the hyperparameters $\mu$ and $\Sigma$ after every iteration, based on the learned values of $\{\phi_k\}$, using Maximum Likelihood estimate. 
$\kappa_1$,$\kappa_2$ can also be updated, but in our implementation we set them to $0.001$ and $0.1$ respectively, based on empirical evaluation on one held-out video. The 
threshold $thres$ was also similarly fixed.

\section{Generative Process for Video Segments}
In the previous section, we considered the entire video as a single block, as the TCCRP PPF for any tracklet $i$ involves $(Z,C)$-values from all the previously seen tracklets 
throughout the video. However, this need not be very accurate, as in a particular part of the video some mixture components (entities) may be more common than anywhere else, and 
for any $i$, $Z_i$ may depend more heavily on the $Z$-values in temporally close tracklets than the ones far away. This is because, a TV-series video consists of \emph{temporal 
segments} like scenes and shots, each characterized by a subset of persons (encoded by binary vector $B_S$). The tracklets attached to a segment $s$ cannot be associated with 
persons not listed by $B_s$. To capture this notion we propose a new model: Temporally Coherent Chinese Restaurant Franchise (TC-CRF) to model a 
video temporally segmented by $S$ (see Section~\ref{sec:def}).

\subsection{Temporally Coherent Chinese Restaurant Franchise}
Chinese Restaurant Process is the PPF associated with Dirichlet Process. Hierarchical Dirichlet Process (HDP)~\cite{hdp} aimed at modelling \emph{grouped data sharing same 
mixture components}. It assumes a group-specific distribution $\pi_s$ for every group $s$. The generative process is:
\begin{small}
\begin{eqnarray}\label{eq:hdp}
 \hat{p}_k \sim Beta(1,\alpha), p_k= \hat{p}_k\prod_{i=1}^{k-1}{(1-\hat{p}_{i-1})}, \phi_k \sim H \textbf{     }\forall k \in [1,\infty) \nonumber \\
 \pi_s \sim p \forall s \in [1,M]; Z_i \sim \pi_{S(i)}, Y_i \sim \mathcal{N}(\phi_{Z_i},\Sigma_1)  \forall i \in [1,N]
\end{eqnarray}
\end{small}
where datapoint $i$ belongs to the group $S(i)$. The PPF corresponding to this process is obtained by marginalizing the distributions $p$ and $\{\pi\}$, and is called the 
\emph{Chinese Restaurant Franchise} process, elaborated in~\cite{hdp}. In our case, we can modify this PPF once again to incorporate TC, analogously to TC-CRP, to have 
Temporally Coherent Chinese Restaurant Franchise (TC-CRF) Process.
In our case, a group corresponds to a temporal segment, and as already mentioned, we want a binary vector $B_s$, which indicates the components that are active in 
segment $s$. 
But HDP assumes that all the components are shared by all the groups, i.e. any particular component can be sampled in any of the groups. We can instead try \emph{sparse modelling} 
by incorporating $\{B_s\}$ into the model, as done in~\cite{FTM} for Focussed Topic Models. For this purpose we put an IBP~\cite{IBP} prior on the $\{B_s\}$ variables, where 
$p(B_{sk}=1|B_{1},\dots,B_{s-1})\propto n_k$ where $n_k$ is the number of times component $k$ has been sampled in all scenes before $s$, and 
$p(B_{sk_{new}}|B_{1},\dots,B_{s-1})\propto \gamma$. The \emph{TC-CRF} PPF is then as follows:
\begin{small}
\begin{eqnarray}\label{eq:PPF3}
 TF(Z_i=k|B_s,Z_{1:i-1},C_{1:i-1},C_i=1) = 0 \textbf{ if } k \in \{Z_{CF(i)}\}-\{0\}& \nonumber \\
                                        = 0 \textbf{ if } B_{sk}=0 & \nonumber \\
                                    \propto \beta  \textbf{ if } k=0 &  \nonumber \\
				    \propto n^{SZC}_{sk1}\textbf{ if } B_{sk}=1, k \in \{Z\}_{s}, k \notin \{Z_{CF(i)}\}& \nonumber \\
                                    \propto \alpha \textbf{ if } B_{sk}=1, k \notin \{Z\}_{s}, k \notin \{Z_{CF(i)}\}
\end{eqnarray}
\end{small}
where $s=S(i)$, the index of the temporal segment to which the datapoint $i$ belongs.
Based on TC-CRF, the generative process of a video, in terms of temporal segments and tracklets, is given below:
\begin{algorithm}[h!]
\caption{TC-CRF Tracklet Generative Process}\label{algo:scgen}
\begin{algorithmic}[1]
\footnotesize
 \STATE $\phi_k \sim \mathcal{N}(\mu,\Sigma_0)$ 
 \FOR{$s=1:M$} 
 \STATE $B_{s} \sim IBP(\gamma,B_1,\dots,B_{s-1})$
 \ENDFOR
 \FOR {$i=1:N$}
 \IF {$S_i=S_{prev(i)}$ }
 \IF {$dist(i,prev(i)) \leq thres$} \STATE $C_i \sim Ber(\kappa1)$
 \ELSE \STATE $C_i \sim Ber(\kappa2)$
 \ENDIF
 \ELSE \STATE $C_i=1$
 \ENDIF
 \IF {$C_i=1$} \STATE draw $Z_i \sim TF(Z_i|B_{S(i)},Z_1,\dots,Z_{i-1},C_1,\dots,C_{i-1},\alpha)$
 \ELSE \STATE $Z_i=Z_{prev(i)}$
 \ENDIF
 \IF {$Z_i=0$} \STATE $Y_i \sim \mathcal{N}(\mu,\Sigma_2)$
 \ELSE \STATE $Y_i \sim \mathcal{N}(\phi_{Z_i},\Sigma_1)$
 \ENDIF
 \ENDFOR
\end{algorithmic}
\end{algorithm}
where $TF$ is the PPF for TC-CRF, and $S(i)$ is the temporal segment index associated with tracklet $i$.

\subsection{Inference}
Inference in TC-CRF can also be performed through Gibbs Sampling. We need to infer the variables $\{B\}$, $\{C\}$, $\{Z\}$ and the components $\{\phi\}$.
In segment $s$, for a datapoint $i$ where $C_i=1$, a component $\phi_k$ may be sampled 
with $p(B_{sk}=1,Z_i=k|B_{-sk},Z_{-i}) \propto n^{SZC}_{sk1}$, which is the number of times $\phi_k$ has been sampled within the same segment. If $\phi_k$ has never 
been sampled within the segment but has been sampled in other segments, $p(B_{sk}=1,Z_i=k|B_{-sk},Z_{-i}) \propto \alpha n_k$, where $n_k$ is the number of segments where 
$\phi_k$ has been sampled (Corresponding to $p(B_{sk})=1$ according to IBP), and $\alpha$ is the CRP parameter for sampling a new component. Finally, a completely new component 
may be sampled with probability proportional to $\alpha$. Note that $p(B_{sk}=0,Z_i=k)=0 \forall k$.

\section{Relationship with existing models}
TC-CRP draws inspirations from several recently proposed Bayesian nonparametric models, but is different from each of them. 
It has three main characterestics: 1) Changepoint-variables $\{C\}$ 2) Temporal Coherence and Spatio-temporal cues  3) Separate component for non-face tracklets. 
The concept of changepoint variable was used in Block-exchangeable Mixture Model~\cite{BE}, which showed that this significantly speeds up the 
inference. But in BEMM, the Bernoulli parameter of changepoint variable $C_i$ depends on $Z_{prev(i)}$ while in TC-CRP it depends on $dist(i,prev(i))$.
Regarding spatio-temporal cues, the concept of providing additional weightage to self-transition was introduced in sticky HDP-HMM~\cite{hdphmm}, but this model does 
not consider change-point variables. Moreover, it uses a transition distribution $P_{k}$ for each mixture component $k$, which increases the  model complexity. Like BEMM~\cite{BE} we 
avoid this step, and hence our PPF (Eq~\ref{eq:PPF1}) does not involve $Z_{prev(i)}$. DDCRP~\cite{ddcrp} defines distances between every pair of datapoints, and associates 
a new datapoint $i$ with one of the previous ones ($1,\dots,i-1$) based on this distance. Here we consider distances between a point $i$ and its predecessor $prev(i)$ only. 
On the other hand, DDCRP is unrelated to the original DP-based CRP, as its PPF does not consider $n^{Z}_{k}$: the number of previous datapoints assigned to component $k$. 
Hence our method is significantly different from DDCRP. Finally, the first two rules of TC-CRP PPF are novel.

TC-CRF is inspired by HDP~\cite{hdp}. However, once again there are three differences mentioned above hold good. In addition,
the PPF of TC-CRF itself is different from Chinese Restaurant Franchise as described in~\cite{hdp}. The original CRF is defined in terms of two concepts: tables and dishes, where 
tables are local to individual restaurants (data groups) while dishes (mixture components) are global, shared across restaurants (groups). Also individual datapoints are assigned 
mixture components indirectly, through an intermediate assignment of tables. The concept of table, which comes due to marginalization of group-specific mixture distributions, 
results in complex book-keeping, and the PPF for datapoints is difficult to define. Here we avoid this problem, by skipping tables and directly assigning mixture components to 
datapoints in Eq~\ref{eq:PPF3}. Inspiration of TC-CRF is also drawn from IBP-Compound Dirichlet Process~\cite{FTM}. But the inference process of~\cite{FTM} is complex, 
since the convolution of the DP-distributed mixture distribution and the sparse binary vector is difficult to marginalize by integration. We avoid this step by directly defining 
the PPF (Eq~\ref{eq:PPF3}) instead of taking the DP route. This approach of directly defining the PPF was taken for DD-CRP~\cite{ddcrp} also.

\section{Experiments on Person Discovery}\label{sec:perdisc}
One particular entity discovery task that has recently received a lot of attention is person discovery from movies/ TV series.
We carried out extensive experiments for person discovery on TV series videos of various lengths. We collected three episodes of The Big Bang Theory (Season 1). Each episode is 
20-22 minutes long, and has 7-8 
characters (occurring in at least 100 frames). We also collected 6 episodes of the famous Indian TV series ``The Mahabharata'' from Youtube. Each episode of this series is 40-45 
minutes long, and have 15-25 prominent characters. So here, each character is an entity. These videos are much longer than those studied in similar works like~\cite{trklnk}, and 
have more characters. Also, these videos are challenging because of the somewhat low quality and motion blur.
Transcripts or labeled training sets are unavailable for all these videos. As usual in the literature~\cite{fcclus}\cite{trklnk}, we represent the persons with their faces. 
We obtained face detections by running the OpenCV Face Detector on each frame separately.
As described in Section~\ref{sec:def} the face detections were all converted to grayscale, scaled down to $30\times30$, and reshaped to form $900$-dimensional vectors.
We considered tracklets of size $R=10$ and discarded smaller ones. The dataset details are given in Table 1.
%

\begin{table}
 \centering
\scriptsize
\begin{tabular}{| c | c | c | c | c | c |}
 \hline
\textbf{Dataset}     & \textbf{\#Frames}      & \textbf{\#Detections}  & \textbf{\#Tracklets}  & \textbf{\#Entities}   & \textbf{Entity Type}\\
\hline
BBTs1e1              &     32248              &     25523              &     2408           &         7       & Person(Face)            \\
BBTs1e3              &     31067              &     21555              &     1985           &         9       & Person(Face)            \\              
BBTs1e4              &     28929              &     20819              &     1921           &         8       & Person(Face)            \\
Maha22               &     66338              &     37445              &     3114           &        14       & Person(Face)            \\
Maha64               &     72657              &     65079              &     5623           &        16	      & Person(Face)	          \\
Maha65               &     68943              &     53468              &     4647           &        22       & Person(Face)            \\
Maha66               &     87202              &     76908              &     6893           &        17	      & Person(Face)	          \\
Maha81               &     78555              &     62755              &     5436           &        22       & Person(Face)            \\
Maha82 	             &     86153              &     52310              &     4262           &        24       & Person(Face)            \\
\hline
\end{tabular}\label{tab:data}
\caption{{Details of datasets}}
\end{table}
\begin{figure}\label{fig:atoms}
 \centering
\includegraphics[width=3.3in,height=0.5in]{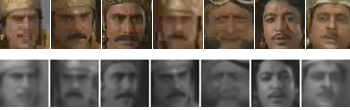} 
\caption{\scriptsize{Face detections (top), and the corresponding atoms (reshaped to square images) found by TC-CRP (bottom)}}
\end{figure}
\begin{figure}
 \centering
\includegraphics[width=3.3in,height=0.3in]{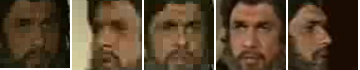} 
\caption{\scriptsize{Different atoms for different poses of same person}}\label{fig:poses}
\end{figure}

\subsection{Alternative Methods}\label{sec:base}
A recent method for face clustering using track information is WBSLRR~\cite{wbslrr} based on Subspace Clustering. Though in~\cite{wbslrr} it is used for clustering detections 
rather than tracklets, the change 
can be made easily. Apart from that, we can use Constrained Clustering as a baseline, and we choose a recent method~\cite{conspec2}. TC and frame conflicts are encoded as 
must-link and don't-link constraints respectively. A big problem is that the number of 
clusters to be formed is unknown. For this purpose, we note that the \emph{tracklet matrix} formed by juxtaposing the tracklet vectors should be \emph{approximately low-rank} 
because of the similarity of spatio-temporally close tracklet vectors. Such representation of a video as a low-rank matrix has been attempted earlier~\cite{rpca}~\cite{denoise}. 
We can find a low-rank representation of the tracklet matrix by any suitable method, and use the rank as the number of clusters to be formed in spectral clustering.
We found that, among these the best performance is given by Sparse Bayesian Matrix Recovery (SBMR)~\cite{sbmr}. Others are either too slow 
(BRPCA~\cite{brpca}), or recover matrices with ranks too low (OPTSPACE~\cite{kesh}) or too high (RPCA~\cite{rpca}). 
Finally, we compare against another well-known sequential BNP method- the sticky HDP-HMM~\cite{hdphmm}.

\subsection{Performance Measures}
The task of entity discovery with all their tracks is novel and complex, and has to be judged by suitable measures.  We discard the clusters that have less than 10 assigned 
tracklets. It turns out that the remaining clusters cover about $85-95\%$  of all the tracklets. Further, there are some clusters which have 
mostly ($70\%$ or more) false (non-entity) tracklets. We discard these from our evaluation. We call the remaining clusters as \emph{significant clusters}.
\emph{We say that a cluster $k$ is ``pure'' if at least $70\%$ of the tracklets assigned to it belong to any one person  $A$}(say Sheldon for a BBT video, or Arjuna for a 
Mahabharata video). 
We also declare that the cluster $k$ and its corresponding mixture component $\phi_k$ corresponds to the person $A$. Also, then $A$ is considered to be \emph{discovered}.
The threshold of purity was set to $70\%$ because we found this roughly the minimum purity needed to ensure that a component mean is visually 
recognizable as the entity (after reshaping to $d\times d$) (See Fig. 4, 5).
We measure the \emph{Purity: fraction of significant clusters that are pure, i.e. correspond to some entity}. We also measure \emph{Entity Coverage: the number of persons (entity) 
with at least 1 cluster (at least 10 tracklets) corresponding to them}. Next, we measure \emph{Tracklet Coverage: the fraction of tracklets that are assigned to pure clusters}. 
Effectively, these tracklets are \emph{discovered}, and the ones assigned to impure clusters are lost. 

\subsection{Results}\label{subsec:res}
The results on the three measures discussed above are shown in Tables 2,3,4. 
In terms of the three measures, TC-CRF is usually the most accurate, followed by TC-CRP, and then sHDP-HMM. This demonstrates that BNP methods are more suitable to the task.
The constrained spectral clustering-based method is competitive on the purity measure, but fares very poorly in terms of tracklet coverage. This is because, it forms many small 
pure clusters, and a few very large impure clusters which cover a huge fraction of the tracklets. Thus, a large number of tracklets are lost. 

It may be noted that the \emph{number of significant clusters formed} is a matter of concern, especially from the user' perspective. A small number of clusters allow him/her to
 get a quick summary of the video. Ideally there should be one cluster per entity, but that is not possible due to the significant appearance variations 
(See Figure~\ref{fig:poses}). The number of clusters formed per video by the different methods is indicated in Table 2. 
It appears that none of the methods have any clear advantage over the others in this regard.
In the above experiments, we used tracklets with size $R=10$. We varied this number and found that, for $R=5$ and even $R=1$ (dealing with detections individually), the 
performance of TC-CRF, TC-CRP and sHDP-HMM did not change significantly. On the other hand, the matrix returned by SBMR had higher rank (120-130 for $R=1$) as the number of 
tracklets increased.

\begin{table}
 \centering
\scriptsize
\begin{tabular}{| c | c | c | c | c | c |}
 \hline
\textbf{Dataset}      &  \textbf{TCCRF}       & \textbf{TCCRP}        & \textbf{sHDPHMM}      & \textbf{SBMR+}       & \textbf{WBSLRR}\\
                      &                       &                       &                       & \textbf{ConsClus}    & \\
\hline
BBTs1e1               &  \textbf{0.88} (48)   & 0.75 (36)             & 0.84 (44)             &      0.67 (48)       &       0.73 (45)  \\
BBTs1e3               &  \textbf{0.88} (50)   & 0.83 (40)             & 0.76 (37)             &      0.80 (15)       &       0.67 (43)  \\
BBTs1e4               &  \textbf{0.93} (40)   & 0.89 (36)             & 0.83 (29)             &      0.77 (31)       &       0.71 (41)  \\
Maha22                &  0.91 (67)            & 0.87 (69)             & 0.86 (74)             & \textbf{0.94} (44)   &       0.83 (79)  \\
Maha64                &  \textbf{0.95} (113)  & 0.92 (105)            & 0.91 (97)             &      0.85 (88)       &       0.75 (81)  \\
Maha65                &  \textbf{0.97} (95)   & 0.89 (85)             & 0.90 (89)             &      0.86 (76)       &       0.82 (84)  \\
Maha66                &  0.91 (76)            & \textbf{0.96} (73)    & 0.95 (80)             &      0.87 (84)       &       0.81 (81)  \\
Maha81                &  \textbf{0.89} (91)   & \textbf{0.89} (88)    & 0.84 (95)             &      0.87 (84)       &       0.74 (78)  \\
Maha82                &  \textbf{0.92} (52)   & 0.88 (50)             & 0.86 (58)             &      0.78 (63)       &       0.83 (64)  \\
\hline
\end{tabular}\label{tab:pur}
\caption{Purity results for different methods. The number of significant clusters are written in brackets}
\end{table}
\begin{table}
 \centering
\scriptsize
\begin{tabular}{| c | c | c | c | c | c |}
 \hline
\textbf{Dataset}     &  \textbf{TCCRF}      & \textbf{TCCRP}       & \textbf{sHDPHMM}     & \textbf{SBMR+}     & \textbf{WBSLRR}\\
                     &                      &                      &                      & \textbf{ConsClus}  & \\
\hline
BBTs1e1              &  \textbf{6}          &  \textbf{6}          &          5           &       5            &       4              \\
BBTs1e3              &  \textbf{9}          &  7                   &          6           &       8            &       7              \\
BBTs1e4              &  6                   &  \textbf{8}	   &   \textbf{8}         &       6            &  \textbf{8}          \\
Maha22               &  \textbf{14}         &  \textbf{14}         &   \textbf{14}        &      10            &  \textbf{14}         \\
Maha64               &  \textbf{14}         &  13                  &   \textbf{14}        &      11            &       13             \\
Maha65               &  17                  &  \textbf{19}         &         17           &      13            &       17             \\
Maha66               &  13                  &  \textbf{15}         &         13           &       9            &       11             \\
Maha81               &  \textbf{21}         &  \textbf{21}         &         20           &      14            &       20             \\
Maha82               &  \textbf{21}         &  19                  &         20           &      10            &       16             \\
\hline
\end{tabular}\label{tab:ccov}
\caption{Entity Coverage results for different methods}
\end{table}
\begin{table}
 \centering
\scriptsize
\begin{tabular}{| c | c | c | c | c | c |}
 \hline
\textbf{Dataset}     &  \textbf{TCCRF}      & \textbf{TCCRP}      & \textbf{sHDPHMM}    & \textbf{SBMR+}   & \textbf{WBSLRR}\\
                     &                      &                     &                     & \textbf{ConsClus}& \\
\hline
BBTs1e1              &    \textbf{0.82}     &    0.67             &  0.79               &      0.29        &    0.73\\
BBTs1e3              &    0.86              &    \textbf{0.88}    &  0.68               &      0.09        &    0.53\\
BBTs1e4              &    \textbf{0.92}     &    0.82             &  0.78               &      0.22        &    0.62\\
Maha22               &    \textbf{0.90}     &    \textbf{0.90}    &  0.86               &      0.43        &    0.69\\
Maha64               &    \textbf{0.93}     &    0.90             &  0.81               &      0.39        &    0.62\\
Maha65               &    \textbf{0.94}     &    0.85             &  0.91               &      0.40        &    0.68\\
Maha66               &    0.74              &    \textbf{0.80}    &  0.68               &      0.43        &    0.65\\
Maha81               &    \textbf{0.80}     &    0.75             &  0.66               &      0.46        &    0.50\\
Maha82               &    0.76              &    \textbf{0.81}    &  0.64               &      0.37        &    0.64\\
\hline
\end{tabular}\label{tab:tcov}
\caption{Tracklet Coverage results for different methods}
\end{table}

\subsection{Online Inference}\label{subsec:olinf}
We wanted to explore the case of streaming videos, where the frames appear sequentially and old frames are not stored.
This is the online version of the problem, the normal Gibbs Sampling will not be possible. For each tracklet $i$, we will have to infer $C_i$ and $Z_i$ based on $C_{prev(i)}$, 
$Z_{prev(i)}$ and the $\{\phi_k\}$-vectors learnt from $\{Y_1,Y_2,\dots,Y_{i-1}\}$. Once again, $(C_i,Z_i)$ is sampled as a block as above, and the term 
$p(Z_i|Z_{-i},C_{i}=a))$ follows from the TC-CRP PPF (Eq \ref{eq:PPF1}). The same thing can be done for TC-CRF also.
Instead of drawing one sample per data-point, an option is to draw several samples and consider 
the mode. In the absence of actual streaming datasets we performed the single-pass inference (Sec~\ref{subsec:olinf}) on two of the videos from each set- Mahabharata and Big Bang 
Theory. We used the same performance measures as above. The existing tracklet clustering methods discussed in Sec~\ref{sec:base} are incapable in the online setting, and sticky 
HDP-HMM is the only alternative. The results are presented in Table 5, which show TC-CRP to be doing the best on the Mahabharata videos and TC-CRF on the Big Bang Theory ones. 
Notably, the figures for TC-CRP and TC-CRF in the online experiment are not significantly lower than those in the offline experiment (except one or two exceptions), unlike 
sHDP-HMM. This indicates that the proposed methods converge quickly, and so are more efficient offline.

\begin{table}
 \centering
\scriptsize
\begin{tabular}{| c || c | c | c |}
 \hline
\textbf{Dataset}              &   \multicolumn{3}{c|}{Maha65}     \\
\hline
\textbf{Measure}              &   TC-CRF      &    TC-CRP              &    sHDPHMM       \\
\hline
Purity                        &   0.86 (56)   &   \textbf{0.89}(79)    &   0.84 (82)      \\
Entity Coverage               &   14          &   15                   &   \textbf{16}    \\
Tracklet Coverage             &   0.75        &   \textbf{0.80}        &   0.77           \\
\hline
 \hline
\textbf{Dataset}              &    \multicolumn{3}{c|}{Maha81}       \\
\hline
\textbf{Measure}              &   TC-CRF     &    TC-CRP         &    sHDPHMM       \\
\hline
Purity                        &   0.71 (55)  & \textbf{0.84}(74) &   0.70(57)  \\
Entity Coverage               &   19         & \textbf{21}       &   17        \\
Tracklet Coverage             &   0.51       & \textbf{0.62}     &   0.49      \\
\hline
 \hline
\textbf{Dataset}              &     \multicolumn{3}{c|}{BBTs1e1}    \\
\hline
\textbf{Measure}              &    TC-CRF      &    TC-CRP    &    sHDPHMM      \\
\hline
Purity                        & \textbf{0.87} (39)  & 0.73 (33)    &  0.50 (14) \\
Entity Coverage               &   \textbf{5}   & 3            &  3              \\
Tracklet Coverage             & \textbf{0.80}  & 0.65         &  0.40           \\
\hline
 \hline
\textbf{Dataset}              &    \multicolumn{3}{c|}{BBTs1e4}    \\
\hline
\textbf{Measure}              &   TC-CRF         &    TC-CRP         &    sHDPHMM       \\
\hline
Purity                        &  \textbf{0.92} (45)   & 0.88 (32)    &  0.75(28)        \\
Entity Coverage               &  \textbf{7}      & 6                 &  \textbf{7}      \\
Tracklet Coverage             &  \textbf{0.87}   & 0.81              &  0.67            \\
\hline
\end{tabular}
\caption{Online (single-pass) analysis on 4 videos}
\end{table}

\subsection{Outlier Detection / Discovery of False Tracklets}
Face Detectors such as~\cite{facedet} are trained on static images, and applied on the videos on per-frame basis. This approach itself has its challenges~\cite{viddet}, and the 
complex videos we consider in our experiments do not help matters. As a result, there is a significant number of \emph{false (non-face) detections}, many of which occur in 
successive frames and hence get linked as tracklets. Identifying such junk tracklets not only helps us to improve the quality of output provided to the users, but may also help 
to adapt the detector to the new domain, by retraining with these new negative examples, as proposed in~\cite{viddet2}.

We make use of the fact 
that false tracklets are relatively less in number (compared to the true ones), and hence at least some of them can be expected to deviate widely from the mean of the tracklet 
vectors. This is taken care of in the TC-CRP tracklet model, through the component $\phi_0$ that has very high variance, and hence is most likely to generate the unusual 
tracklets. We set this variance $\Sigma_2$ as $\Sigma_2=c\Sigma_1$, where $c>1$. The tracklets assigned $Z_i=0$ are reported to be junk by our model. It is expected that 
high $c$ will result in lower recall but higher precision (as only the most unusual tracklets will go to this cluster), and low $c$ will have the opposite effect. 
We study this effect on two of our videos- Maha65 and Maha81 (randomly chosen) in Table 6 (See Fig. 7 for illustration).
As baseline, we consider K-means or spectral clustering of the tracklet vectors. We may expect that one of the smaller clusters 
should contain mostly the junk tracklets, since faces are roughly similar (even if from different persons) and should be grouped together. However, for different values of 
$K$ (2 to 10) we find that the clusters are roughly of the same size, and the non-face tracklets are spread out quite evenly. Results are reported for the best $K$ 
($K=10$ for both).
Note that because of the large number of tracklets (Table I) it is difficult to count the total number of non-face ones. So for measuring
 \emph{recall}, we simply mention the \emph{number of non-face tracklets recovered (recall*)}, instead of the \emph{fraction}. It is clear that TC-CRP significantly outperforms 
clustering on both precision and recall*.

\begin{figure}\label{fig:false}
 \centering
\includegraphics[width=3.3in,height=0.5in]{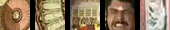} 
\caption{\small{Non-face tracklet vectors (reshaped) recovered by TC-CRP. Note that one face tracklet has been wrongly reported as non-face}}
\end{figure}

\begin{table}
 \centering
\scriptsize
\begin{tabular}{| c | c | c | c | c |}
 \hline
\textbf{Dataset}              &     \multicolumn{2}{c|}{Maha65}     &    \multicolumn{2}{c|}{Maha81}    \\
\hline
\textbf{Method}               &       Precision    &    Recall*    &       Precision        &       Recall* \\
\hline
KMeans                        &        0.22        &      73       &        0.19            &        39 \\
Constrained Spectral          &        0.30        &      12       &        0.12            &        16 \\
TCCRP (c=5)                   &        0.98        &      79       &        0.57            &        36 \\
TCCRP (c=4)                   &        0.98        &      87       &        0.64            &        47 \\
TCCRP (c=3)                   &        0.95        &      88       &        0.62            &        54 \\
TCCRP (c=2)                   &        0.88        &     106       &        0.50            &        57 \\
\hline
\end{tabular}
\caption{Discovery of non-face tracklets}
\end{table}

\subsection{Evaluation of TC enforcement}
The aim of TC-CRP and TC-CRF is to encourage TC at the semantic level, that spatio-temporally close but non-overlapping tracklets should belong to the same entity. In the 
Bayesian models like sHDP-HMM, TC-CRP and TC-CRF, these cues are modelled with probability distributions, in WBSLRR with convex regularization and in constrained clustering they 
are encoded as hard constraints. We now evaluate how well the different methods have been able to enforce these cues. We create ground-truth tracks by linking the tracklets which 
are spatio-temporally close to each other (with respect to the chosen threshold $thres$ in the generative process), and belong to the same entity. All the tracklets in each 
ground-truth track should be assigned to the same cluster. This is the task of \emph{tracklet linking}.
We measure \emph{what fraction of the these ground-truth tracks have been assigned entirely to single clusters} by 
the different methods. We do not compare SBMR+ConsClus, since it uses hard constraints. The results are shown in Table 7. We find that TC-CRF is the best once again, followed 
by TC-CRP and sHDP-HMM. WBSLRR has significantly poorer performance, though it springs a surprise on BBTs1e1.

\begin{table}
 \centering
\scriptsize
\begin{tabular}{| c | c | c | c | c |}
 \hline
\textbf{Dataset}     &  \textbf{TCCRF}      & \textbf{TCCRP}      & \textbf{sHDPHMM}    & \textbf{WBSLRR}\\
                     &                      &                     &                     & \\
\hline
BBTs1e1              &    0.65              &    0.54             &  0.42               &    \textbf{0.93}\\
BBTs1e3              &    \textbf{0.74}     &    0.71             &  0.59               &    0.22\\
BBTs1e4              &    \textbf{0.72}     &    0.69             &  0.54               &    0.34\\
Maha22               &    \textbf{0.83}     &    0.81             &  0.80               &    0.61\\
Maha64               &    \textbf{0.80}     &    \textbf{0.80}    &  0.79               &    0.55\\
Maha65               &    \textbf{0.86}     &    0.81             &  0.81               &    0.63\\
Maha66               &    \textbf{0.86}     &    0.79             &  0.78               &    0.52\\
Maha81               &    \textbf{0.86}     &    0.82             &  0.83               &    0.61\\
Maha82               &    \textbf{0.89}     &    0.86             &  0.84               &    0.64\\
\hline
\end{tabular}
\caption{Fraction of ground truth tracks that are fully linked}
\end{table}

\begin{table}
 \centering
\scriptsize
\begin{tabular}{| c | c | c | c | c |}
 \hline
\textbf{Dataset}      & \textbf{TCCRP}        & \textbf{sHDPHMM}      & \textbf{SBMR+}       & \textbf{WBSLRR}\\
                      &                       &                       & \textbf{ConsClus}    & \\
\hline
Cars                  &  0.94 (35)            & 0.92 (12)             & \textbf{1.00} (54)   &       0.24 (21)  \\
Aeroplanes            &  \textbf{0.95} (43)   & 0.87 (15)             &      0.84 (44)       &       0.21 (24)  \\
\hline
\hline
\end{tabular}
 \centering
\scriptsize
\begin{tabular}{| c | c | c | c | c |}
 \hline
\textbf{Dataset}     & \textbf{TCCRP}       & \textbf{sHDPHMM}     & \textbf{SBMR+}     & \textbf{WBSLRR}\\
                     &                      &                      & \textbf{ConsClus}  & \\
\hline
Cars                 &  \textbf{5}          &   \textbf{5}         &   \textbf{5}       &        2             \\
Aeroplanes           &  \textbf{6}          &          5           &   \textbf{6}       &        4             \\
\hline
\hline
\end{tabular}
 \centering
\scriptsize
\begin{tabular}{| c | c | c | c | c |}
 \hline
\textbf{Dataset}     & \textbf{TCCRP}      & \textbf{sHDPHMM}    & \textbf{SBMR+}   & \textbf{WBSLRR}\\
                     &                     &                     & \textbf{ConsClus}& \\
\hline
Cars                 &    0.73             &  0.69               & \textbf{1.00}    &    0.04\\
Aeroplanes           &    \textbf{0.93}    &  0.70               &      0.88        &    0.09\\
\hline
\end{tabular}
\caption{Purity, Entity Coverage and Tracklet Coverage results for different methods on Cars and Aeroplanes videos}
\end{table}

\begin{figure}
 \centering
\includegraphics[width=3.3in,height=0.45in]{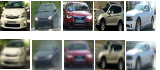} 
\caption{\scriptsize{Car detections (top), and the corresponding atoms (reshaped to square images) found by TC-CRP (bottom)}}
\end{figure}

\section{Discovery of Non-person Entities}
To emphasize the fact that our methods are not restricted to faces or persons, we used two short videos-one of cars and another of aeroplanes. The cars video consisted 
of 5 cars of different colors, while the aeroplanes video had 6 planes of different colors/shapes. These were created by concatenating shots of different cars/planes in the 
Youtube Objects datasets~\cite{yobj}. The objects were detected using the Object-specific detectors~\cite{felzen}. Since here the color is the chief
distinguishing factor, we scaled the detections down to $30\times 30$ and reshaped them separately in the 3 color channels to get 2700-dimensional vectors. Here $R=1$ was used,
 as these videos are much shorter, and using long tracklets would have made the number of data-points too low. Both videos have 750 frames. The \emph{Cars} video has 694 
detections, and the \emph{Aeroplanes} video has 939 detections. The results are shown in Table 8. Once again, TC-CRP does well.

\section{Semantic Video Summarization}
In this section, we discuss how the above results on entity discovery can be used to obtain a sematic summary of the video. For this purpose we 
consider two approaches: entity-based and shot-based.

\subsection{Entity-based Summarization}
The process of entity discovery via tracklet clustering results in formation of clusters.In case of the Bayesian methods like TC-CRF, TC-CRP and sHDP-HMM, each cluster can be 
represented by the mean vector of the corresponding mixture component. In case of non-Bayesian approaches like SBMR+Consclus and WBSLRR, it is possible to compute the cluster 
centers as the mean of the tracklet vectors assigned to each cluster. Each cluster vector $\phi_k$ can be reshaped to form a visual representation of the cluster.
 This representation of clusters provides us a visual list of the entities present in the video, which is what we call \emph{entity-based summarization} of the video.

Any summary should have two properties: 1) It should be \emph{concise} 2) It should be \emph{representative}. Along these lines, an entity-based summary should have the property 
that it should cover as many entities as possible, with least number of clusters. On the other hand, the selected clusters should cover a sufficiently large fraction of all the 
tracklets. In our evaluation of entity discovery (Section~\ref{sec:perdisc}) we have measured \emph{Entity Coverage}, \emph{Tracklet Coverage} and \emph{Number of significant 
clusters}. These same measures are useful in evaluating the summarization. Entity Coverage and Tracklet Coverage should be high, and number of significant clusters should be low
(See Figures 8,9). To make the evaluation more comprehensive, we define two more measures: 
1) \emph{Conciseness:} defined as the ratio of Entity Coverage to the number of significant clusters, and 2) \emph{Representativeness:} defined as the ratio of the Tracklet 
Coverage to the number of significant clusters.

The results are shown in the Tables 9,10. We find that in terms of Conciseness, TC-CRP turns out to be the best, while the other methods are all comparable when averaged across the 
videos. In terms of Representativeness, TC-CRP is once again the best by a long way, while TC-CRF and sHDP-HMM are at par. The non-Bayesian methods are way behind.

\begin{table}
 \centering
\scriptsize
\begin{tabular}{| c | c | c | c | c | c |}
 \hline
\textbf{Dataset}     &  \textbf{TCCRF}      & \textbf{TCCRP}      & \textbf{sHDPHMM}    & \textbf{SBMR+}   & \textbf{WBSLRR}\\
                     &                      &                     &                     & \textbf{ConsClus}& \\
\hline
BBTs1e1              &    0.13              &    \textbf{0.17}    &  0.11               &      0.10        &    0.09\\
BBTs1e3              &    0.18              &    0.18             &  0.16               &   \textbf{0.53}  &    0.16\\
BBTs1e4              &    0.15              &    0.22             &  \textbf{0.28}      &      0.19        &    0.20\\
Maha22               &    0.21              &    0.20             &  0.19               &   \textbf{0.23}  &    0.18\\
Maha64               &    0.12              &    0.12             &  0.14               &      0.11        &    \textbf{0.16}\\
Maha65               &    0.18              &    \textbf{0.22}    &  0.19               &      0.17        &    0.20\\
Maha66               &    0.17              &    \textbf{0.21}    &  0.16               &      0.11        &    0.14\\
Maha81               &    0.23              &    0.24             &  0.21               &      0.17        &    \textbf{0.26}\\
Maha82               &    \textbf{0.40}     &    0.38             &  0.34               &      0.16        &    0.25\\
\hline
\end{tabular}
\caption{Conciseness results for different methods}
\end{table}

\begin{table}
 \centering
\scriptsize
\begin{tabular}{| c | c | c | c | c | c |}
 \hline
\textbf{Dataset}     &  \textbf{TCCRF}      & \textbf{TCCRP}      & \textbf{sHDPHMM}    & \textbf{SBMR+}   & \textbf{WBSLRR}\\
                     &                      &                     &                     & \textbf{ConsClus}& \\
\hline
BBTs1e1              &    1.7               &    \textbf{1.86}    &  1.80               &      0.60        &    1.60\\
BBTs1e3              &    1.72              &    \textbf{2.2}     &  1.83               &      0.60        &    1.23\\
BBTs1e4              &    2.3               &    2.28             &  \textbf{2.69}      &      0.71        &    1.51\\
Maha22               &    \textbf{1.39}     &    1.30             &  1.16               &      0.99        &    0.87\\
Maha64               &    0.82              &    \textbf{0.86}    &  0.84               &      0.44        &    0.76\\
Maha65               &    0.99              &    1.00             &  \textbf{1.02}      &      0.53        &    0.81\\
Maha66               &    0.97              &    \textbf{1.10}    &  0.85               &      0.51        &    0.80\\
Maha81               &    \textbf{0.88}     &    0.85             &  0.69               &      0.55        &    0.64\\
Maha82               &    1.46              &    \textbf{1.62}    &  1.10               &      0.59        &    1.00\\
\hline
\end{tabular}
\caption{Representativeness ($\times 100$) results for different methods}
\end{table}

\subsection{Shot-based Summarization}
Another way of summarization is by a collection of \emph{shots}. ~\cite{csum} follows this approach, and selects a subset of the shots based on the total number of characters 
(entities), number of prominent characters (entities) etc. A shot~\cite{shotseg} is a contiguous sequence of frames that consist of the same set of entities. It is possible to 
organize the video 
into temporal segments based on the cluster indices assigned to the tracklets. In a frame $f$, let $\{Z\}_f$ denote the set of cluster labels assigned to the tracklets that cover
 frame $f$. For two successive frames $f1$ and $f2$, if $\{Z\}_{f1}=\{Z\}_{f2}$ we say that they belong to the same temporal segment, i.e. $T(f2)=T(f1)$. But if 
$\{Z\}_{f1} \neq \{Z\}_{f2}$, then
 we start a new temporal segment, i.e. $T(f2)=T(f1)+1$. By this process, the frames of the video are partitioned into temporal segments. The cluster labels are supposed to 
correspond to entities, so each temporal segment should correspond to a shot. Each such segment can be easily represented with any one frame, since all the frames in a segment 
contain the same entities. This provides us a \emph{shot-based summarization} of the video.

As in the case with entities, once again a large number of temporal segments are created by this process, with several adjacent segments corresponding to the same set of 
entities. This happens because often several clusters are formed for the same entity. Analogous to Entity Coverage, we define \emph{Shot Coverage} as the total number of 
true shots that have at least one temporal segment lying within it. We then define \emph{significant segments} as those which cover a sufficient number (say 100) of frames. 
Finally, we define \emph{Frame Coverage} as the fraction of the frames which come under the significant segments. 

To evaluate such shot-based summarization, once again we need to consider the two basic properties: conciseness and representativeness. These are measured in exact analogy to the 
entity-based summarization discussed above (See Figures 10,11). The \emph{Conciseness} of the summary is defined as the ratio of the Shot Coverage to the number of 
significant segments, while the \emph{Representativeness} of the summary is defined as the ratio of the Frame Coverage to the number of significant segments. The results are shown
 in Tables 11 and 12. This time we find that in terms of representativeness TC-CRF leads the way, followed by TC-CRP. In terms of conciseness the best performance is given by 
WBSLRR, which however does poorly in terms of representativeness.

\begin{table}
 \centering
\scriptsize
\begin{tabular}{| c | c | c | c | c | c |}
 \hline
\textbf{Dataset}     &  \textbf{TCCRF}      & \textbf{TCCRP}      & \textbf{sHDPHMM}    &\textbf{SBMR+}   & \textbf{WBSLRR}\\
                     &                      &                     &                     &\textbf{ConsClus}& \\
\hline
BBTs1e1              &    \textbf{0.86}     &  0.80               &  0.75               &  0.74           & 0.80\\
BBTs1e3              &    \textbf{0.84}     &  0.74               &  0.71               &  0.82           & 0.64\\
BBTs1e4              &    0.67              &  0.57               &  0.55               &  \textbf{0.75}  & 0.60\\
Maha22               &    0.41              &  0.39               &  0.40               &  0.30           & \textbf{0.53}\\
Maha64               &    0.32              &  \textbf{0.34}      &  \textbf{0.34}      &  0.26           & 0.27\\
Maha65               &    0.30              &  0.29               &  0.30               &  0.24           & \textbf{0.32}\\
Maha66               &    0.14              &  0.14               &  0.12               &  0.11           & \textbf{0.21}\\
Maha81               &    0.37              &  0.34               &  0.36               &  0.24           & 0.17\\
Maha82               &    0.36              &  0.33               &  0.38               &  0.23           & \textbf{0.41}\\
\hline
\end{tabular}
\caption{Conciseness results for different methods}
\end{table}

\begin{table}
 \centering
\scriptsize
\begin{tabular}{| c | c | c | c | c | c |}
 \hline
\textbf{Dataset}     &  \textbf{TCCRF}      & \textbf{TCCRP}      & \textbf{sHDPHMM}    & \textbf{SBMR+}   & \textbf{WBSLRR}\\
                     &                      &                     &                     & \textbf{ConsClus}& \\
\hline
BBTs1e1              &    0.72              &    0.73             &  0.75               &  0.77            & \textbf{0.80}\\
BBTs1e3              &    0.68              &    0.63             &  0.64               &  0.68            & \textbf{0.74}\\
BBTs1e4              &    0.88              &  \textbf{0.93}      &  0.89               &  0.75            & 0.87\\
Maha22               &  \textbf{0.66}       &    0.61             &  0.63               &  0.53            & 0.24\\
Maha64               &  \textbf{0.42}       &    0.41             &  0.40               &  \textbf{0.42}   & 0.23\\
Maha65               &  \textbf{0.47}       &    0.43             &  0.45               &  0.18            & 0.26\\
Maha66               &    0.43              &    0.42             &  0.42               &  \textbf{0.44}   & 0.10\\
Maha81               &    0.45              &  \textbf{0.46}      &  \textbf{0.46}      &  0.19            & 0.29\\
Maha82               &  \textbf{0.59}       &  \textbf{0.59}      &  0.57               &  0.38            & 0.24\\
\hline
\end{tabular}
\caption{Representativeness ($\times 100$) results for different methods}
\end{table}

\begin{figure}
 \centering
\includegraphics[width=3.3in,height=1.6in]{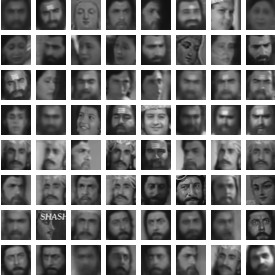}
\caption{\small{Entity-based summarization of Mahabharata Episode 22 using TC-CRF. Each image is a reshaped cluster mean.}}
\includegraphics[width=3.3in,height=2.0in]{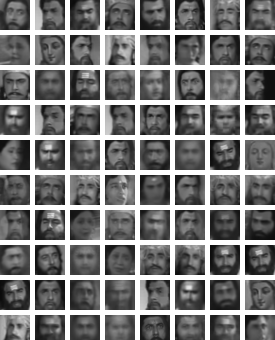}
\caption{\small{Entity-based summarization of Mahabharata Episode 22 using WBSLRR. WBSLRR creates many more clusters than TC-CRF, but both discover the same number of persons (14). 
Hence the summary by TC-CRF is more concise.}}
\label{fig:cons1}
\end{figure}
\begin{figure}
 \centering
\includegraphics[width=3.3in,height=1.5in]{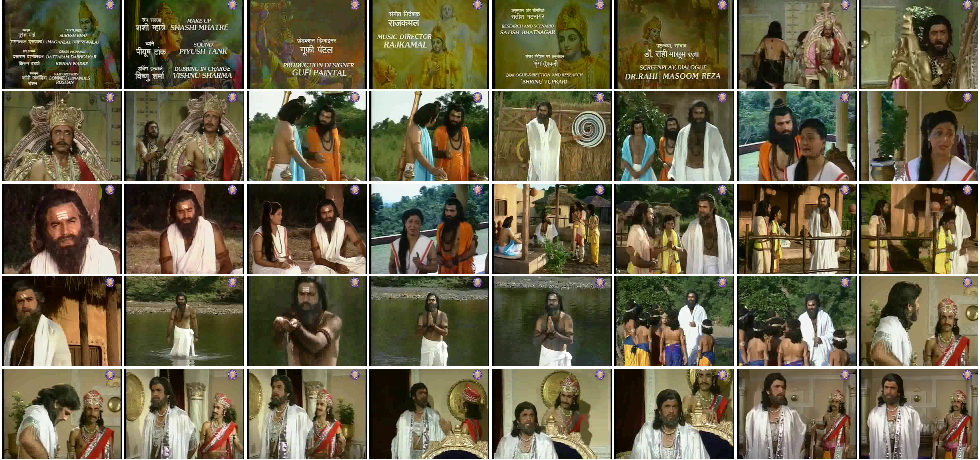}
\caption{\small{Shot-based summarization of Mahabharata Episode 22 using TC-CRF. Each image is a keyframe from a significant segment.}}
\includegraphics[width=3.3in,height=2.0in]{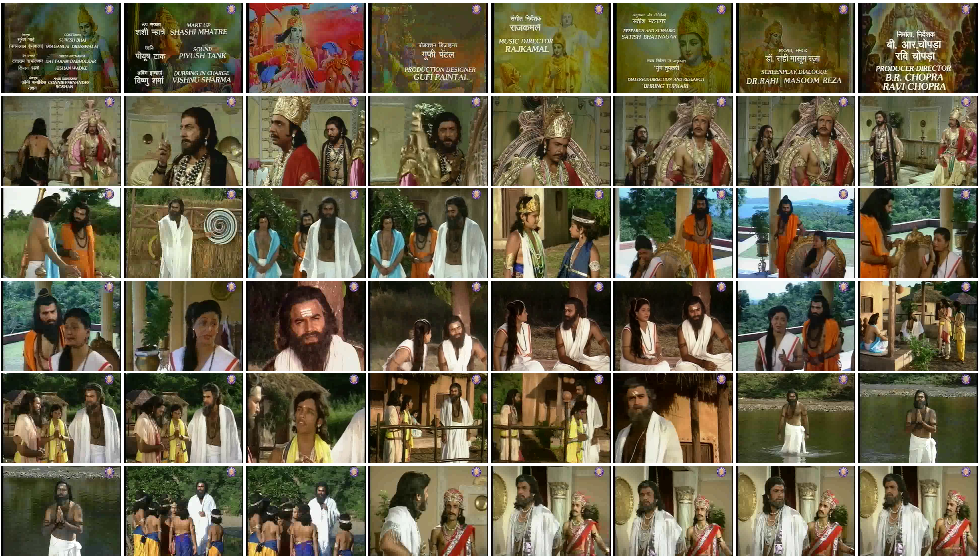} 
\caption{\small{Shot-based summarization of Mahabharata Episode 22 using SBMR+ConsClus. SBMR+ConsClus creates more significant segments to cover roughly the same set of 
true shots as TC-CRF, so TC-CRF summary is more concise}}
\label{fig:cons2}
\end{figure}

\section{Conclusion}
In this paper, we considered an entity-driven approach to video modelling. We represented videos as sequences of tracklets, with each tracklet associated with an entity. 
We defined entity discovery as the task of discovering the repeatedly appearing entities in the videos, along with all their appearances, and cast this as tracklet clustering.
We considered a Bayesian nonparametric approach to tracklet clustering which can automatically discover the number of clusters to be formed. We leveraged the Temporal Coherence 
property of videos to improve the clustering by our first model: TC-CRP. The second model TC-CRF was a natural extension to TC-CRP, to jointly model short temporal segments within
 a video, and further improve entity discovery. These methods were empirically shown to have several additional abilities like performing online entity discovery efficiently, and 
detecting false tracklets. Finally, we used these results for semantic video summarization in terms of the discovered entities.

\end{document}